\documentclass[10pt,twocolumn,letterpaper,pagenumbers]{article}

\usepackage[pagenumbers]{iccv} 
\usepackage{multirow} 
\usepackage{kotex}

\definecolor{iccvblue}{rgb}{0.21,0.49,0.74}
\usepackage[pagebackref,breaklinks,colorlinks,allcolors=iccvblue]{hyperref}

\title{Enhancing Generalization in Data-free Quantization via Mixup-class Prompting}

\author{
Jiwoong Park\textsuperscript{1}\thanks{Equal contribution.}\hspace{0.1em} \thanks{Work done while at Hanyang University.} \quad
Chaeun Lee\textsuperscript{2}\footnotemark[1]\hspace{0.1em} \thanks{Work done while at Rebellions.} \quad
Yongseok Choi\textsuperscript{3} \quad
Sein Park\textsuperscript{4} \quad
Deokki Hong\textsuperscript{4} \quad
Jungwook Choi\textsuperscript{5}\thanks{Corresponding author.} \\
\\[0.01em]
\textsuperscript{1}HyperAccel \quad
\textsuperscript{2}KRAFTON \quad
\textsuperscript{3}SK Telecom \quad
\textsuperscript{4}Rebellions \quad
\textsuperscript{5}Hanyang University \\
{\tt\small \textsuperscript{1}jiwoong.park@hyperaccel.ai, \textsuperscript{2}chaeun.lee@krafton.com, \textsuperscript{1}yongseokchoi@sk.com} \\
{\tt\small \textsuperscript{4}\{seinpark, dk.hong\}@rebellions.ai, \textsuperscript{5}choij@hanyang.ac.kr}
}

\begin{document}
\maketitle
\begin{abstract}
Post-training quantization (PTQ) improves efficiency but struggles with limited calibration data, especially under privacy constraints.
Data-free quantization (DFQ) mitigates this by generating synthetic images using generative models such as generative adversarial networks (GANs) and text-conditioned latent diffusion models (LDMs), while applying existing PTQ algorithms.
However, the relationship between generated synthetic images and the generalizability of the quantized model during PTQ remains underexplored.
Without investigating this relationship, synthetic images generated by previous prompt engineering methods based on single-class prompts suffer from issues such as polysemy, leading to performance degradation.

We propose \textbf{mixup-class prompt}, a mixup-based text prompting strategy that fuses multiple class labels at the text prompt level to generate diverse, robust synthetic data.
This approach enhances generalization, and improves optimization stability in PTQ.
We provide quantitative insights through gradient norm and generalization error analysis.
Experiments on convolutional neural networks (CNNs) and vision transformers (ViTs) show that our method consistently outperforms state-of-the-art DFQ methods like GenQ. Furthermore, it pushes the performance boundary in extremely low-bit scenarios, achieving new state-of-the-art accuracy in challenging 2-bit weight, 4-bit activation (W2A4) quantization. Our code is available at \href{https://github.com/aiha-lab/Mixup-class-Prompting}{https://github.com/aiha-lab/Mixup-class-Prompting}
\end{abstract}    
\section{Introduction}
\label{sec:intro}

Deploying deep learning models efficiently is crucial for real-world applications, especially in resource-constrained environments. Quantization is a widely used technique that reduces model size and computational cost by converting high-precision weights and activations into lower-bit representations. Among various quantization techniques, post-training quantization (PTQ) is particularly attractive because it does not require retraining the model, unlike quantization-aware training (QAT). However, PTQ typically relies on small calibration data, making it prone to overfitting and accuracy degradation, particularly in data-sensitive scenarios where access to training data is restricted~\cite{pham2024metaaug, wei2022qdrop}.

To address this challenge, data-free quantization (DFQ) has emerged as a promising alternative, eliminating the need for real calibration data by generating synthetic data. Recent DFQ methods leverage generative models, such as generative adversarial networks (GANs) and latent diffusion models (LDMs), to create synthetic data. One of the state-of-the-art approaches, GenQ~\cite{li2025genq}, utilizes pre-trained text-conditioned LDMs~\cite{rombach2022high} to generate images based on class labels, achieving state-of-the-art DFQ performance.

Although the pre-trained text-conditioned LDM enables the generation of synthetic images for PTQ without requiring additional training of generative models, the effectiveness of these synthetic images for PTQ remains under-explored. The limited semantic information within simple class labels often leads to issues such as polysemy, which degrades PTQ performance by creating a mismatch between synthetic and real data distributions. For example, the word “kite” may refer to either a toy or a bird, and this ambiguity could result in synthetic images that do not faithfully represent the intended class. This issue of polysemy-induced, out-of-distribution data is particularly detrimental in the context of PTQ. Unlike model training, PTQ relies on a very small calibration set (e.g., 1024 images in our setup) to determine quantization parameters for the entire model. In this extreme low-data regime, even a small fraction of corrupted samples can severely skew the statistical moments (e.g., min/max values) used for calibration, leading to substantial and unrecoverable accuracy degradation.

To systematically address this, a concrete metric is needed to quantify how different prompting strategies affect the final quantized model's ability to generalize. To this end, we propose using the theoretical relationship between the generalization gap and the empirical gradient norm of quantization parameters to quantitatively assess the effectiveness of different prompting strategies. This framework posits that a larger gradient norm during calibration implies a looser generalization bound, signaling a higher risk of overfitting to the synthetic data. Using this metric, we observe that existing single-class prompting methods indeed produce a large gradient norm, indicating poor generalization.

To address this generalization challenge, we propose \textbf{mixup-class prompt}: a simple yet effective text prompting strategy for DFQ. Instead of generating images using single-class text prompts, we combine two class labels within a single prompt (e.g., “a realistic photo of a kite and a vulture”), encouraging the LDM to generate images that contain mixed-class characteristics. This approach mitigates the impact of polysemous labels by blending ambiguous objects into broader contexts and increases data diversity, consequently improving the generalization of the quantized model.

Through extensive experiments on both convolutional neural networks (CNNs) and vision transformers (ViTs), we demonstrate that our approach consistently outperforms existing DFQ methods across various architectures and bit-width settings. 
Our method achieves state-of-the-art accuracy in low-bit quantization settings and exhibits substantial improvements over prior works, particularly in ViT-based models.
Our main contributions are as follows:
\begin{itemize}
    \item We introduce \textbf{mixup-class prompt}, a simple yet effective mixup-based text prompting to reduce generalization gap, resulting in mitigating polysemy issues in generating synthetic images for PTQ.
    \item We provide analytical insights into why mixup prompting improves PTQ by analyzing its impact on optimization stability through generalization gap.
    \item We conduct extensive empirical evaluations on multiple CNN and ViT architectures, demonstrating that our method achieves superior DFQ performance compared to existing approaches in extremely low-bit scenarios.
    \item We compare our method against both standard data augmentation techniques and alternative prompting strategies, showing its effectiveness in improving synthetic data quality for PTQ.
\end{itemize}

\section{Backgrounds}

\subsection{Quantization: PTQ and DFQ}
PTQ is a technique used to improve the efficiency of deep learning models by reducing bit-precision of weights and activations, thereby cutting down model size and computational cost~\cite{li2021brecq, wei2022qdrop, lee2023flexround}. 
This differs from QAT as it does not require retraining the model with whole training data. 
However, PTQ typically relies on a small amount of calibration dataset, making it susceptible to overfitting and accuracy degradation~\cite{liu2023pd, pham2024metaaug}.

To address this challenge, DFQ is considered a promising alternative, eliminating the need for real calibration data by generating synthetic data. 
Earlier research proposed training GANs to produce synthetic data, enabling effective quantization without access to the training dataset.
Recent DFQ methods leverage generative models, such as text-conditioned LDMs, to create synthetic data. 
GenQ~\cite{li2025genq} leverages pre-trained text-conditioned LDMs to generate images based on class labels, achieving state-of-the-art performance.

\subsection{Data Augmentation and Quantization}

Data augmentation is widely used technique to generate new samples by modifying existing data, thereby enhancing the generalization performance of deep learning models. 
Various data augmentation techniques have been proposed, among which Mixup~\cite{zhang2018mixupempiricalriskminimization}, CutMix~\cite{yun2019cutmix}, and ResizeMix~\cite{qin2020resizemix} have gained significant attention.
Mixup generates new training samples by linearly interpolating pairs of examples and their labels in pixel space, effectively smoothing decision boundaries and enhancing generalization.
CutMix augments data by cutting and pasting image patches between samples, allowing mixed regions to reflect a combination of labels, which improves feature learning and reduces overfitting.
ResizeMix extends this idea by merging resized patches from different images, aiming to preserve object information while maintaining the regularization benefits of label mixing.

In the domain of quantization, applying these augmentation techniques is a key area of research. 
It is particularly crucial in DFQ for alleviating the significant challenge of limited diversity in synthetic data. 
This lack of diversity could lead to overfitting, a problem exacerbated in PTQ due to the sensitivity of small calibration datasets.
For instance, MixMix~\cite{li2021mixmix} has improved performance by applying image mixing during image generation. 
Furthermore, MetaAug~\cite{pham2024metaaug} is the first work that has demonstrated the efficacy of data augmentation in PTQ and showed performance increase. 
However, the effect of data augmentation on quantization remains largely underexplored, especially for mixup-based data augmentation.

\subsection{Synthetic Data Generation with LDM}

Diffusion models are probabilistic generative models that iteratively add noise to a prior distribution and train a model to reverse this process for denoising. 
Once trained, they generate synthetic images by sampling random Gaussian noise and mapping it back to the prior distribution. 
The success of large-scale text-conditioned LDMs~\cite{rombach2022high} has driven their widespread adoption in synthetic dataset generation and augmentation. 
Several studies have explored text-based prompting for dataset synthesis in training from scratch, self-supervised learning, quantization, and transfer learning~\cite{sariyildiz2023fake, fan2024scaling, he2022synthetic, li2025genq, tian2024learning}. 
Strategies include appending hypernyms (e.g., ``tench, cyprinid''), background elements (e.g., ``tench in river'')~\cite{sariyildiz2023fake}, while some approaches use language models to generate captions from text labels~\cite{he2022synthetic, tian2024learning}. 
Additionally, several studies~\cite{wang2024enhance, islam2024diffusemix} propose LDM-based mixup augmentation. 
\section{Observations and Methodologies}

\subsection{Challenges in PTQ with Synthetic Generation}
\label{subsec:challenge}

Synthetic images generated by text-conditioned LDMs are influenced by the semantic meaning of the text prompt, which guides the denoising process.
In other words, the object depicted in the synthetic image is determined by the single-class text label used to represent the target image class within the prompt.
For instance, a synthetic image could be generated with ”[template] [C]”, coined as \textbf{single-class prompt}, where "[template]" is selected from pre-defined template pool and "[C]" corresponds to a class text label from a given set.

If the prompt provides rich contextual information or a detailed description of the target class, the generated image is more likely to contain the correct object.
However, in the absence of sufficient context, the semantic ambiguity of the text can cause the model to generate an image that does not include the intended object—an issue known as polysemy~\cite{white2022schr}.
For instance, in ImageNet~\cite{russakovsky2015imagenet}, the class, “kite" (n01608432), refers to a species of bird, but the word, "kite", has both meanings of kites as toys and birds. 
If the text-to-image model is biased toward generating the toy interpretation, synthetic dataset as calibration dataset includes samples from out-of-distribution classes for ImageNet~\cite{russakovsky2015imagenet}. 

\label{sec:formatting}
\begin{figure}[t!]
\includegraphics[width=8.5cm]{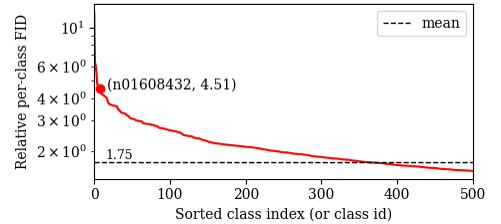}
\caption{Top 50\% classes of ImageNet sorted by RPC-FID. A class with larger score (e.g.,  "kite" (n01608432)) indicates that synthetic and real training images are more likely separated.}
\label{fig::fig_fid}
\vspace{-3mm}
\end{figure}

To quantify the polysemy issue, we propose a metric called \textit{Relative Per-Class Fr\'echet Inception Score (RPC-FID)},  which captures the extent to which individual classes in the synthetic dataset are affected by semantic ambiguity:
\begin{equation} \label{eqn::rpc-fid}
\text{RPC-FID} = \frac{\text{FID}((m_{1}^{real}, C_{1}^{real}),(m_{1}^{syn}, C_{1}^{syn}))}{\text{FID}((m_{2}^{real}, C_{2}^{real}),(m_{3}^{real}, C_{3}^{real}))} ,
\end{equation}
where $m_{i}^{real}$ and $C_{i}^{real}$ are mean vector and covariance matrix of a set of feature vectors, $f(X_{i}^{real})$, extracted from a pre-trained model (Inception-v3), $f$, for an disjoint subset of samples from the distribution of real training dataset, $X^{real}$;
i.e., $m_{i}^{real} = \textbf{E}[X_{i}^{real}]$ and $C_{i}^{real} = \textbf{Var}[X_{i}^{real}]$. 
$m_{i}^{syn}$ and $C_{i}^{syn}$ correspond to those of synthetic dataset~\cite{heusel2017gans}.

\begin{figure*}[t]
\centering
\includegraphics[width=0.8\linewidth]{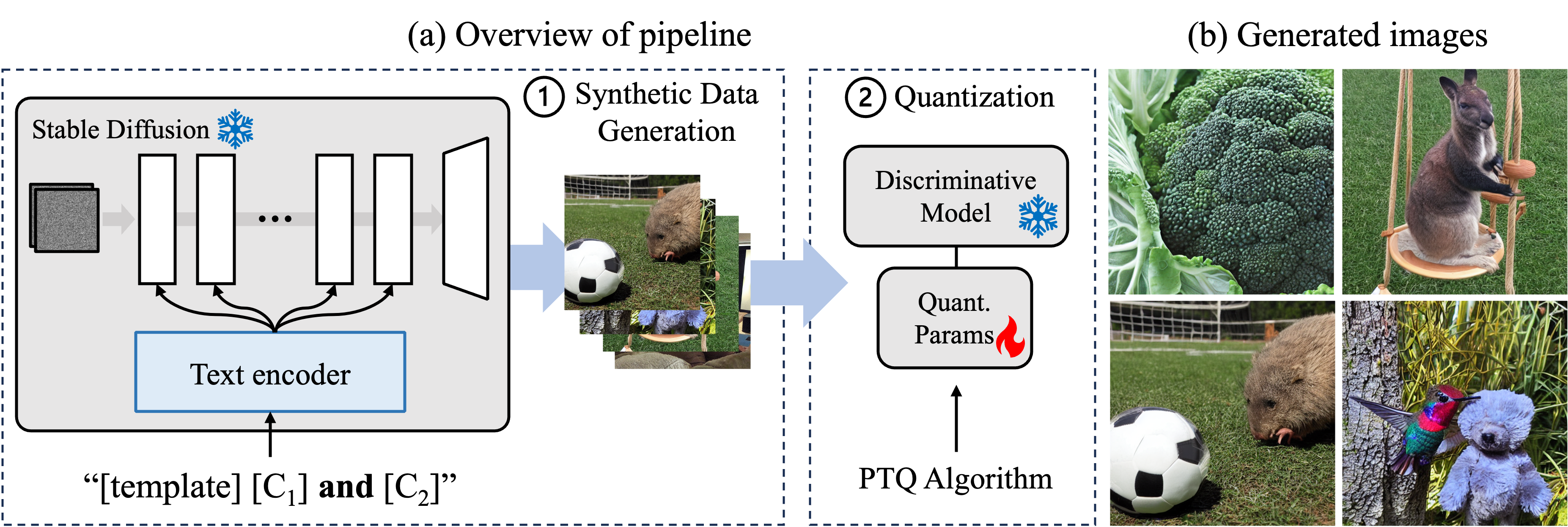}
\caption{(a) The overview of pipeline for our method and (b) examples of generated images using \textbf{mixup-class prompt}; (a) text prompts based on \textbf{mixup-class prompt} are conditioned on LDMs (Stable Diffusion) to generate synthetic images. At the following step, with the generated images as calibration dataset, quantization parameters are optimized using PTQ algorithms. (b) Synthetic images in the calibration dataset are generated from \textbf{mixup-class prompt}.}
\vspace{-5mm}
\label{fig::fig_pipeline_ex}
\end{figure*}

In Eq.~\ref{eqn::rpc-fid}, the numerator represents FID score of a sampled subset of synthetic dataset and that of real training dataset, while the denominator corresponds to the FID score between two disjoint subsets of the real training dataset.
This metric estimates the relative deviation of synthetic dataset from real training dataset considering the inherent intra-class variance of disjoint sampled real training dataset. 

RPC-FID reveals that synthetic images generated by text-conditioned LDMs suffer from out-of-distribution classes, where most of recent works~\cite{sariyildiz2023fake, he2022synthetic, li2025genq, fan2024scaling} generate images using the variants of \textbf{single-class prompt}.
Fig.~\ref{fig::fig_fid} shows RPC-FID scores of top 50\% ImageNet classes where synthetic image is generated with \textbf{single-class prompt}~\cite{li2025genq}. 
For instance, the class, "kite" (n01608432), shows a higher RPC-FID score because most of synthetic images for this class depict kite as toys rather than birds, even though "kite" class in ImageNet refers to birds. 
Although GenQ~\cite{li2025genq} attempts to mitigate this issue by filtering synthetic images based on model confidence scores and statistical properties of the quantized model, this approach requires extra synthetic images and incurs computation overhead.

Although RPC-FID is introduced to quantify how faithfully synthetic images capture the polysemous nature of the real training data, its implications for the adverse effects of synthetic images on the optimization of quantization parameters during PTQ remain underexplored.
In a more general sense, there is a need for a metric that evaluates which dataset is most suitable for optimizing quantization parameters.
This need becomes particularly important in the context of PTQ, where optimization is performed within a parameter space that is typically smaller than that of the original model weights, while relying on a significantly smaller calibration dataset than the original training set.
Such conditions make PTQ more susceptible to overfitting and convergence to poor local optima~\cite{wei2022qdrop, pham2024metaaug}, underscoring the importance of a concrete and quantifiable criterion.

To quantify this effect, we introduce recent theoretical research to demystify the connection between the generalization gap and empirical gradient norm of model parameters, $\theta$, during model training process~\cite{li2019generalization, an2021can}. 
For a training dataset denoted as $\mathcal{X}=\{x_1, x_2, \cdots, x_{N}\}$ and validation/test dataset $\{y_1, y_2, \cdots, y_{M}\}$, the empirical generalization gap $\hat{\mathcal{G}}_{M}$ is estimated as
\begin{equation} \label{eqn::generalization_gap}
\hat{\mathcal{G}}_{M}(\theta) = \frac{1}{M} \sum_{i=1}^{M}{\mathcal{L}(\theta;y_{i})} - \frac{1}{N} \sum_{i=1}^{N}{\mathcal{L}(\theta;x_{i})},    
\end{equation}
where $\mathcal{L}$ indicates the loss and $\lim_{M \rightarrow \infty }{\hat{\mathcal{G}}_{M}} = \mathcal{G}$, which indicates that the empirical generalization gap converges to generalization gap.

The generalization gap is proved to be upper bounded to the function of the empirical gradient norm:
\begin{equation}\label{eqn:error_gap}
\mathcal{G}(\theta_T) = O\Bigg(\dfrac{1}{N}\sqrt{E_{x\sim \mathcal{X}}\big[\sum_{t=1}^{T}{\dfrac{\gamma_{t}^2}{\sigma_{t}^2}\mathbf{g}(t)}\big]}\Bigg),       
\end{equation}
where $\mathbf{g}(t) = E_{\theta_{t-1}}\big[\dfrac{1}{n}\sum_{i=1}^{n}{||\nabla\mathcal{L}(\theta_{t-1}, x_{i})||_{2}^{2}}\big]$ represents the gradient $L_2$ norm for a mini-batch of size $n$ at the $t^{th}$ iteration of model $\theta_t$, while $\gamma_t$ and $\sigma_t$ denote the learning rate and the standard deviation of Gaussian noise in the stochastic gradient descent process.

In the aspect of PTQ, without loss of generality, we restrict the scope of model parameters, $\theta$, to the quantization parameters while the original model parameters are frozen and training dataset to calibration dataset.
For calibration datasets, $\mathcal{X}_{1}, \mathcal{X}_{2}, \cdots$, if the upper bound in Eq.~\ref{eqn:error_gap} is smaller than others, then the generalization gap is expected to be smaller. 
Accordingly, for the optimal calibration dataset, $\mathcal{X}^{*}$, the generalization gap is minimized, which is the expected behavior for the well-trained full-precision model.

\subsection{Mixup-Class Prompt}

If there are sufficient real data resources to compose multiple candidates for calibration dataset, it is possible to find the optimal set of data indices that could reduce generalization gap in Eq.~\ref{eqn:error_gap}.
However, with the deficiency of real data resouces or when no available real data is available, it becomes difficult to compse a set of calibration dataset candidates.
Instead, text-conditioned LDMs could provide calibration dataset candidates by varying the strategy for text prompting based on \textbf{single-class prompt} which uses one class text label and a template, for instance, from a subset of CLIP templates~\cite{li2025genq}.
Several works~\cite{sariyildiz2023fake} have proposed prompt strategies based on \textbf{single-class prompt}---for example, by appending extra background text to diversify image background, or by appending the definition of the class text label to clarify the meaning of the class text label.

\begin{figure}[t!]
  \centering
  \centerline{\hspace*{-5mm}\includegraphics[width=8cm]{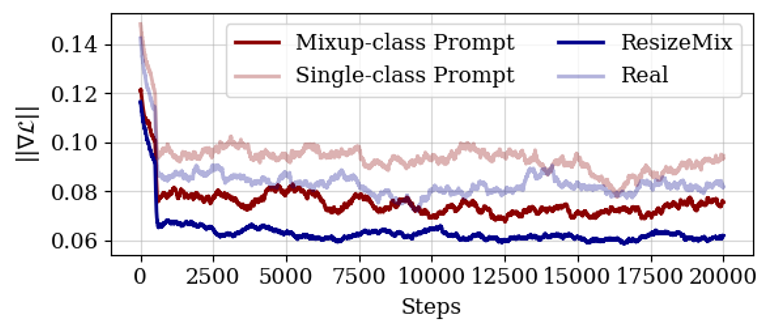}}
  \vspace{-1mm}
  \centerline{(a) Activation scale}\medskip
  \vspace{-2mm}
  \centerline{\hspace*{-5mm}\includegraphics[width=8cm]{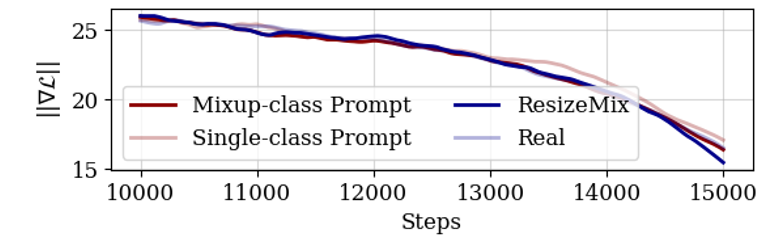}}
  \vspace{-1mm}
  \centerline{(b) Weight rounding}\medskip
  \vspace{-2mm}
  \centerline{\hspace*{-5mm}\includegraphics[width=8cm]{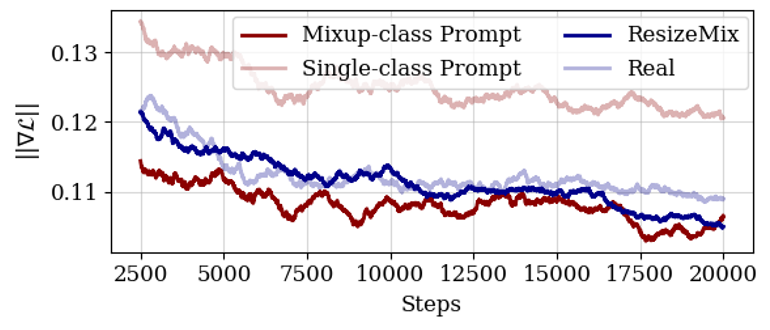}}
  \vspace{-1mm}
  \centerline{(c) Weight scale}\medskip
  \vspace{-2mm}
  \caption{Comparison of the trace of gradient norms with respect to quantization parameters: weight/activation scale and weight rounding.Each curve is derived from different image augmentation or generation methods.}
  \label{fig::fig_gradnorm}
\end{figure}

As shown in Fig.~\ref{fig::fig_gradnorm}, the empirical gradient norm with respect to quantization parameters for \textbf{single-class prompt} is larger than that of real data, which indicates that the upper bound for the generalization gap in Eq.~\ref{eqn:error_gap} becomes higher.
Motivated by classical machine learning approaches to reduce the generalization gap in Eq.~\ref{eqn:error_gap}, we focus on introducing regulization effect through text prompt engineering. 
Among various regularization methods, we propose leveraging mixup~\cite{zhang2018mixupempiricalriskminimization} at the text prompt level, coined as \textbf{mixup-class prompt}, based on the following motivations: 
First, mixup is one of powerful regularization methods to soften decision boundaries by mixing multiple objects in the image, which leads to a reduced graident norm~\cite{zhang2018mixupempiricalriskminimization}. 
Secondly, implementing mixup at text prompt level is straightforward and does not require additional text pool such as a text pool to describe background images beyond a set of templates and class text labels. 

As illustrated in Fig.~\ref{fig::fig_pipeline_ex} (a), \textbf{mixup-class prompt} implement mixup in the form of ``[template] [C\textsubscript{1}] and [C\textsubscript{2}]'', where [template] is randomly selected from a subset of CLIP templates proposed in GenQ \cite{li2025genq}.
[C\textsubscript{1}] and [C\textsubscript{2}] are randomly sampled class text labels from the training dataset. 
After the synthetic image generation step, quantization parameters for the pre-trained discriminative model are calibrated with existing PTQ algorithms using the synthetic dataset.
For instance, when the discriminative model adopts a CNN architecture, state-of-the-art PTQ algorithms such as Genie-M~\cite{jeon2023genie} could be integrated. 
This method allows two classes to be seamlessly combined, as illustrated in Fig.~\ref{fig::fig_pipeline_ex} (b). 
Since the approach requires no architectural modifications, it is widely applicable across discriminative models and eliminates the need for additional hyperparameter tuning.

\section{Analysis}

As discussed in Eq.~\ref{eqn:error_gap}, the trace of empirical gradient norm with respect to quantization parameters serves as an indicator of the generalizability of the quantized model during PTQ; 
that is, the optimal calibration dataset, $\mathcal{X}^{*}$, exhibits a consistently lower gradient norm across optimization steps.
Fig.~\ref{fig::fig_gradnorm} shows the gradient norm trace of quantization parameters at ResNet50’s layer 2.3 across different calibration dataset, with Genie-M~\cite{jeon2023genie} adopted as PTQ algorithm. 
For \textbf{single-class prompt}, gradient norms exceed those observed with real images, indicating a larger generalization gap and a higher risk of convergence to poor local optima. 
For models prone to overfitting to limited size of calibration dataset, mixup-based data augmentation is known to enhance robustness and provide a regularization effect~\cite{zhang2018mixupempiricalriskminimization}. 
In line with this, applying ResizeMix~\cite{qin2020resizemix} for real images reduces the gradient norms of activation and weight scales, effectively lowering the upper bound of the generalization gap. 
However, like other mixup-based augmentations, ResizeMix requires the specification of hyperparameters such as the mixing ratio, introducing additional optimization complexity.
For \textbf{mixup-class prompt}, it achieves a lower gradient norm compared to that of \textbf{single-class prompt}. 
Furthermore, its gradient norm is even lower than that of real images, leading to a reduced upper bound on the generalization gap and consequently improving the stability of PTQ optimization process.
Unlike ResizeMix~\cite{qin2020resizemix}, as it is free from hyperparameter optimization, it does not require additional efforts.

\begin{figure}[!t]
\begin{minipage}[b]{.49\linewidth}
  \centering
  \centerline{\includegraphics[width=4.3cm]{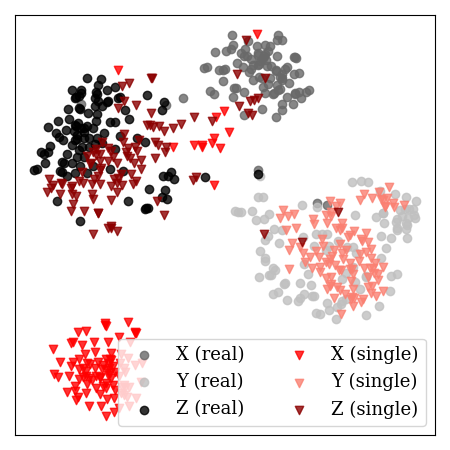}}
  \centerline{(a) real and single-class}\medskip
\end{minipage}
\begin{minipage}[b]{.49\linewidth}
  \centering
  \centerline{\includegraphics[width=4.3cm]{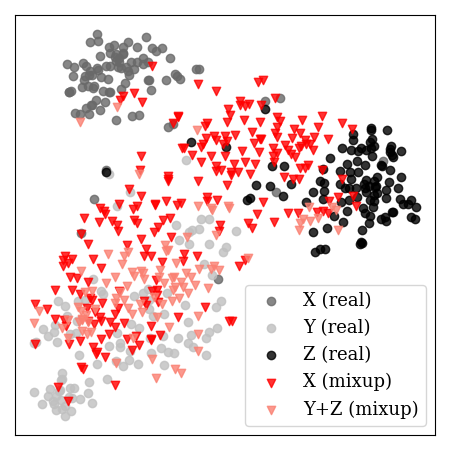}}
  \centerline{(b) real and mixup-class}\medskip
\end{minipage}
\caption{t-SNE plots for the feature map of ResNet50. Samples are drawn from three classes: X stands for ``n01608432'' (kite), Y for ``n01614925'' (bald eagle), and Z for ``n01616318'' (vulture). For (b), X (mixup) indicates both X+Y and X+Z.}
\label{fig::fig_tsne}
\vspace{-5mm}
\end{figure}

There could exist multiple qualitative interpretation on the effect of \textbf{mixup-class prompt} in reducing the upper bound of the generalization gap. 
Among them, one possible explanation is its ability to mitigate polysemy issue which leads to the emergence of out-of-distribution classes, as described in the Sec.~\ref{subsec:challenge}. 
Fig.~\ref{fig::fig_tsne} presents t-SNE visualizations of synthetic images generated using \textbf{single-class prompt} and \textbf{mixup-class prompt}, alongside real data for three selected classes (birds), whose RPC-FID are 4.51, 1.58, and 0.92, respectively. 
For polysemic classes such as “kite,” which shows higher RPC-FID score, the synthetic images forms a distinct cluster by semantic confusion, as shown in Fig.~\ref{fig::fig_tsne} (a). 
In other words, most of synthetic images for the "kite" class depicts kites as toys, rather than birds, which are the intended semantic category in the ImageNet dataset.
Only a few samples in the synthetic dataset represent kites as birds, as suggested by the small number of outlier points deviating from the cluster in Fig.~\ref{fig::fig_tsne} (a). 

After applying \textbf{mixup-class prompt}, Fig.~\ref{fig::fig_tsne} (b) shows improved alignment of feature distributions.
The distinct cluster in Fig.~\ref{fig::fig_tsne} (a) disappears as the "kite" class is mixed with two other bird classes: "bald eagle" and "vulture."
This improvement occurs not by explicitly correcting the polysemous meaning of “kite,” but rather by blending polysemous objects into the background context of less ambiguous classes, which leads to mitigating polysemy issue. 
Although text-conditioned LDMs generate toy-like kites, they are blended within images containing other objects from classes with lower RPC-FID and less semantic ambiguity.
For these two classes, generated images exhibit diverse features, as kites appear as background objects, thereby increasing intra-class diversity.

\begin{table}[t]
\centering
\resizebox{\columnwidth}{!}{%
\begin{tabular}{cccccc}
\hline
\multirow{2}{*}{\textbf{\begin{tabular}[c]{@{}c@{}}Bit-\\ width\end{tabular}}} & \multirow{2}{*}{\textbf{Method}} & \textbf{RN20} & \textbf{RN18} & \textbf{RN50} & \textbf{MBv2} \\
 &  & \textbf{(CIFAR-100)} & \multicolumn{3}{c}{\textbf{(ImageNet)}} \\ \hline
\multicolumn{2}{c}{Baseline} & 70.33 & 71.01 & 76.63 & 72.62 \\ \hline
\multirow{7}{*}{W4A4} & Real & 69.06 & 69.75 & 75.64 & 69.24 \\ \cline{2-6}
 & Qimera & 65.10 & 63.84 & 66.25 & 61.62 \\
 & AdaDFQ & 66.81 & 66.53 & 68.38 & 65.41 \\
 & Genie & 68.35 & 69.66 & 75.59 & 68.54 \\
 & SADAG & \textbf{69.11} & 69.72 & \textbf{75.70} & 68.54 \\
 & GenQ & - & 69.77 & 75.50 & 68.96 \\
 & Ours & \textbf{69.11} & \textbf{69.81} & 75.61 & \textbf{69.07} \\ \hline
\multirow{7}{*}{W3A3} & Real & 65.88 & 67.51 & 72.85 & 59.65 \\ \cline{2-6}
 & Qimera & 46.13 & 1.17 & - & - \\
 & AdaDFQ & 52.74 & 38.10 & 17.63 & 28.99 \\
 & Genie & 65.25 & 66.89 & 72.54 & 55.31 \\
 & SADAG & 65.94 & 67.10 & 72.62 & 56.02 \\
 & GenQ & - & - & - & - \\
 & Ours & \textbf{65.96} & \textbf{67.26} & \textbf{72.76} & \textbf{58.65} \\ \hline
\multirow{5}{*}{W2A4} & Real & - & 66.05 & 70.88 & 56.11 \\ \cline{2-6}
 & Genie & - & 65.10 & 69.99 & 51.47 \\
 & SADAG & - & 65.25 & 70.52 & 51.89 \\
 & GenQ & - & 65.72 & 70.35 & 54.82 \\
 & Ours & - & \textbf{65.90} & \textbf{70.78} & \textbf{55.08} \\ \hline
\end{tabular}
}
\caption{Top-1 accuracy of CNNs on CIFAR-100 and ImageNet. RN corresponds to ResNet and MBv2 to MobileNetV2. `-' indicates data from the baseline papers is unavailable.}
\vspace{-3mm}
\label{tab:cnn}
\end{table}
\begin{table*}[ht!]
\centering
\resizebox{0.95\textwidth}{!}{%
\setlength{\tabcolsep}{15pt}
\begin{tabular}{cccccccc}
\hline
\textbf{Bit-width} & \textbf{Method} & \textbf{ViT-S} & \textbf{ViT-B} & \textbf{DeiT-S} & \textbf{DeiT-B} & \textbf{Swin-S} & \textbf{Swin-B} \\ 
\hline
\multicolumn{2}{c}{Baseline} & 81.39 & 84.54 & 79.85 & 81.80 & 83.23 & 85.27 \\ 
\hline
\multirow{5}{*}{W4A4} 
 & Real & 65.05 & 68.48 & 69.03 & 75.61 & 79.45 & 78.32 \\ \cline{2-8}
 & PSAQ-ViT & - & 60.18 & - & 74.69 & - & 54.21 \\
 & GenQ & - & 67.50 & - & \textbf{76.10} & - & 70.08 \\
 & MimiQ & 55.69 & 62.91 & 62.72 & 74.10 & 70.46 & 73.49 \\
 & Ours & \textbf{64.84} & \textbf{67.52} & \textbf{69.60} & 75.80 & \textbf{79.65} & \textbf{73.53} \\ 
\hline
\multirow{3}{*}{W5A5} 
 & Real & 78.48 & 81.89 & 77.16 & 80.15 & 81.93 & 82.97 \\ \cline{2-8}
 & MimiQ & 70.02 & 78.09 & 72.59 & 78.20 & 80.75 & 82.05 \\
 & Ours & \textbf{78.36} & \textbf{81.97} & \textbf{77.16} & \textbf{79.97} & \textbf{82.03} & \textbf{82.55} \\ 
\hline
\multirow{3}{*}{W8A8}
 & Real & 81.25 & 84.41 & 79.69 & 81.76 & 83.08 & 85.19 \\ \cline{2-8}
 & MimiQ & \textbf{81.30} & \textbf{85.17} & 79.68 & 81.84 & \textbf{83.05} & 84.73 \\
 & Ours & 81.23 & 84.32 & \textbf{79.78} & \textbf{81.85} & \textbf{83.05} & \textbf{85.02} \\ 
\hline
\end{tabular}%
}
\caption{Top-1 accuracy on ImageNet with Vision Transformers.}
\vspace{-2mm}
\label{tab:vit-imagenet}
\end{table*}
\section{Experiments}

\subsection{Experimental Setting}
We evaluate the accuracy of our method on both CNN and ViT models. For CNNs, we validate ResNet20, ResNet18, ResNet50~\cite{he2016deep}, and MobileNetV2~\cite{sandler2018mobilenetv2}. We quantize all CNN models with Genie-M~\cite{jeon2023genie} algorithm and evaluate them under various bit-width settings. The baselines for comparison include Qimera~\cite{choi2021qimera}, AdaDFQ~\cite{qian2023adaptive}, Genie~\cite{jeon2023genie}, SADAG~\cite{dungsharpness}, and GenQ~\cite{li2025genq}. For ViTs, we validate ViT~\cite{dosovitskiy2020image}, Swin~\cite{liu2021swin}, and DeiT \cite{touvron2021training}, including their small and base versions. We quantize these models with RePQ-ViT~\cite{li2023repq}. The baselines for comparison include PSAQ-ViT~\cite{li2022patch} (cited from GenQ), GenQ and MimiQ~\cite{choi2024mimiq}. To ensure a fair comparison with previous methods, we generate 1024 images using Stable Diffusion v1-5 with a guidance scale of 3.5 and 50 denoising steps. Wx/Ay denotes that the weights and activations are quantized to x-bit and y-bit, respectively.

\subsection{Results on CNN and ViTs}
We conduct experiments on two datasets in CNN models, CIFAR-100~\cite{krizhevsky2009learning} and ImageNet~\cite{russakovsky2015imagenet}, where CIFAR-100 represents a relatively easy one, and ImageNet a more challenging one. Table~\ref{tab:cnn} provides the experimental results on CIFAR-100 and ImageNet, comparing our proposed method with existing approaches. CIFAR-100 models in Table~\ref{tab:cnn}, our method achieves comparable accuracy to SADAG in W4A4 and outperforms SADAG by 0.02\% under the W3A3 setting. Improvement in accuracy on ImageNet is more significant compared to CIFAR-100. As shown for the ImageNet models in Table~\ref{tab:cnn}, overall, our method outperforms existing approaches across various models and bit-width. Even in more challenging scenarios such as W3A3 and W2A4, our approach maintains a clear advantage over previous methods, demonstrating its robustness and effectiveness in low bit-width.

We observed more significant improvements in performance when applying our proposed method to ViTs. Table~\ref{tab:vit-imagenet} presents a comparative analysis against previous methods on the ImageNet dataset for various Vision Transformer architectures. Similar to the results obtained with CNNs, the proposed approach consistently outperforms existing techniques. Under the W8A8 setting, it achieves accuracy comparable to that obtained using real data. Furthermore, under low bit-width settings, it demonstrates superior accuracy. Specifically, in the W5A5 setting, it outperforms MimiQ across all models, with a notable accuracy improvement of 8.34\% on ViT-S. In W4A4, the method shows an accuracy gain of approximately $7\sim9\%$ on smaller models such as ViT-S, DeiT-S, and Swin-S, and nearly 5\% on ViT-B.
\section{Ablation study}

\subsection{Comparison with Other Augmentation}
\begin{table}[]
\centering
\resizebox{0.9\columnwidth}{!}{%
\small
\begin{tabular}{cccc}
\hline
\textbf{Augmentation} & \textbf{RN18} & \textbf{RN50} & \textbf{MBv2} \\ \hline
Baseline & 71.01 & 76.63 & 72.62 \\ \hline
No Augmentation & 65.45 & 70.01 & 51.62 \\
Cutmix & 65.48 & 70.06 & 52.10 \\
Mixup & 65.53 & 70.15 & 51.79 \\
ResizeMix & 65.59 & 70.43 & 52.29 \\
MetaAug & 65.20 & 69.74 & 53.32 \\
Mixup-class (Ours) & \textbf{65.90} & \textbf{70.78} & \textbf{55.08} \\ \hline
\end{tabular}%
}

\caption{Accuracy across different augmentation methods under W2A4 using synthetic images as the calibration set.}
\vspace{-2mm}
\label{tab:ablation-aug}
\end{table}
Our research meticulously compared the efficacy of our proposed methodology against established image augmentation techniques, including CutMix, Mixup, ResizeMix, and MetaAug. 
To ensure a rigorous and equitable assessment, these conventional augmentation strategies were applied to images synthesized from \textbf{single-class prompt}.
As evidenced in Table~\ref{tab:ablation-aug}, our method consistently outperformed existing augmentation methods, achieving superior accuracy for all cases. 
This clearly highlights the superiority of our mixup-based augmentation method at the level of text prompts for text-conditioned LDMs over traditional pixel-level methods. 
By generating diverse and semantically meaningful augmentations, our method consistently maintains model accuracy, even under challenging scenario.

\subsection{Impact of Prompt Engineering}
\begin{table}[t!]
\centering
\setlength{\tabcolsep}{6pt}
\resizebox{\columnwidth}{!}{%
\begin{tabular}{cccc}
\hline
\textbf{Prompt} & \textbf{RN18} & \textbf{RN50} & \textbf{MBv2} \\ \hline
Baseline & 71.00 & 76.63 & 72.62 \\ \hline
[template] [C] \textit{hyp} & 65.53 & 70.55 & 54.65 \\
{[template] [C] \textit{def}} & 65.69 & 70.42 & 54.99 \\
{[template] [C] \textit{hyp inside back}} & 65.69 & 70.49 & 54.43 \\
LM generated & 65.72 & 70.48 & 54.33 \\
{[template] [C$_{1}$] and [C$_{2}$] (Ours)} & \textbf{65.90} & \textbf{70.78} & \textbf{55.08} \\ \hline
\end{tabular}%
}
\caption{Comparison of W2A4 accuracy across different prompting methods.}
\vspace{-3mm}
\label{tab:ablation-prompting}
\end{table}
Furthermore, we rigorously assessed our methodology's performance against various prompting methods. 
Table~\ref{tab:ablation-prompting} presents ImageNet test accuracy for models quantized to W2A4, where models are quantized with synthetic dataset generated via established prompting methods; 
'hyp' denotes prompts augmented with a hypernym of the class~\cite{sariyildiz2023fake}; 
'def' indicates prompts incorporating the class definition from WordNet~\cite{miller1994wordnet}; 
'hyp inside b' signifies prompts encompassing both a hypernym and contextual background information from a pre-defined text pool~\cite{sariyildiz2023fake}; 
and 'T5' refers to prompts generated using a T5-base model to formulate sentences containing the class~\cite{he2022synthetic}. 
These comparative results establish that our method exhibits markedly superior to existing prompting methods. 
These results indicate that our prompting method, despite its simplicity, consistently shows robust accuracy under low-precision quantization.

\subsection{Other Diffusion Models}
\begin{table}[]

\resizebox{\columnwidth}{!}{%
\begin{tabular}{c|cc|ccc}
\hline
\multirow{2}{*}{\begin{tabular}[c]{@{}c@{}}\textbf{LDM}\\ \textbf{Models}\end{tabular}}         & \textbf{Bit-width}                & \textbf{Prompt}    & \textbf{RN18}       & \textbf{RN50}       & \textbf{MBv2}    \\ \cline{2-6} 
                                                                              & \multicolumn{2}{c|}{Baseline} & 71.00          & 76.63          & 72.62          \\ \hline
\multirow{4}{*}{\begin{tabular}[c]{@{}c@{}}SD v2\end{tabular}} & \multirow{2}{*}{W4A4}    & mixup     & \textbf{69.85} & \textbf{75.66} & \textbf{69.13} \\
                                                                              &                          & single    & 69.73          & 75.64          & 68.72          \\
                                                                              & \multirow{2}{*}{W2A4}    & mixup     & \textbf{65.93} & \textbf{70.87} & \textbf{54.70} \\
                                                                              &                          & single    & 65.78          & 70.15          & 53.79          \\ \hline
\multirow{4}{*}{\begin{tabular}[c]{@{}c@{}}SD v3\end{tabular}} & \multirow{2}{*}{W4A4}    & mixup     & \textbf{69.94} & \textbf{75.67} & \textbf{69.25} \\
                                                                              &                          & single    & 69.91          & 75.63          & 68.70          \\
                                                                              & \multirow{2}{*}{W2A4}    & mixup     & \textbf{65.94} & \textbf{70.12} & \textbf{53.46} \\
                                                                              &                          & single    & 65.70          & 69.72          & 52.91          \\ \hline
\end{tabular}%
}
\caption{Accuracy comparison across various Stable Diffusion models and discriminative models.}
\vspace{-5mm}
\label{tab:ablation-ldm-models}
\end{table}
To further substantiate the generalizability and robustness of our proposed augmentation methodology, we conducted additional experiments evaluating its efficacy across different text-conditioned LDMs beyond the primary Stable Diffusion v1-5 (SD v1-5) employed in our main study. As shown in the table, we evaluated the accuracy of models quantized to W4A4 and W2A4, using images generated with \textbf{mixup-class} and \textbf{single-class prompt}. The results consistently indicate that the \textbf{mixup-class prompt} yields superior accuracy compared to \textbf{single-class prompt} across two LDMs which have different  architectures, Stable Diffusion v2 (SD v2)~\cite{rombach2022high} and Stable Diffusion v3 (SD v3)~\cite{esser2024scaling}, and across various discriminative models such as ResNet18, ResNet50, and MobileNetV2.

As shown in Table \ref{tab:ablation-ldm-models}, for SD v2, \textbf{mixup-class prompt} consistently achieved higher accuracy than \textbf{single-class prompt} at both W4A4 and W2A4. 
A similar trend is observed with SD v3, where our \textbf{mixup-class prompt} consistently surpassed \textbf{single-class prompt} accuracy. 
This consistent improvement, irrespective of the underlying LDM variant, supports our assertion that the proposed augmentation method operates orthogonally to the LDM architectures. 
This implies that our strategy effectively enhances data augmentation by leveraging the generative capabilities of LDMs, without being intrinsically tied to a particular model architectures. 
This demonstrates the broad applicability and modularity of our approach, suggesting its potential to be seamlessly integrated with future LDM architectures.

\begin{table}[]
\centering
\resizebox{0.9\columnwidth}{!}{%
\begin{tabular}{ccccc}
\hline
\textbf{Bit-width}             & \begin{tabular}[c]{@{}c@{}}\textbf{Embedding}\\ \textbf{Similarity}\end{tabular} & \textbf{RN18}           & \textbf{RN50}           & \textbf{MBv2}           \\ \hline
\multicolumn{2}{c}{Baseline}                                                    & 71.00          & 76.63          & 72.62          \\ \hline
\multirow{3}{*}{W2A4} & High                                                           & 65.71          & 70.51          & 54.34          \\
                      & Low                                                            & 65.71 & 70.55 & 54.54 \\
                      & Random                                                         & \textbf{65.90} & \textbf{70.78} & \textbf{55.08} \\ \hline
                      &                                                                &                &                &               
\end{tabular}%
}
\vspace{-5mm}
\caption{Comparison of \textbf{mixup-class prompt} with varying class embedding similarities. Pairing random class outperforms both high and low similarity-based pairs across all models.}
\label{tab:ablation-label-sim}
\end{table}
\subsection{Semantic Similarity between Class Labels}

For more in-depth study, we investigate the impact of semantic relationships between classes used in our method. 
We conducted an additional ablation study by varying the embedding similarity between class pairs. 
We computed similarity scores following the method proposed in~\cite{Hong_2025_WACV}, which is based on class-representative vectors extracted from the final fully connected layer of a CNN models.

Specifically, under the W2A4 setting, we compared three strategies for selecting the second class in a \textbf{mixup-prompt class}: High similarity, Low similarity, and Random. 
As shown in the table, the High and Low similarity strategies showed comparable performance. 
However, the Random strategy consistently achieved the best accuracy across all downstream models.

These results suggest that enforcing predefined semantic relationships—whether similar or dissimilar—between mixed classes does not necessarily enhance the effectiveness of data augmentation. 
In contrast, the diversity introduced by random pairing appears to offer greater benefits for robustness and generalization, particularly in PTQ. 
This unexpected finding implies that while embeddings may capture semantic meaning, the most effective pairing strategies tend to prioritize diversity over strict semantic alignment. 
These insights highlight the subtlety of prompt engineering, suggesting that maximizing or minimizing semantic similarity is not always the most effective strategy.

\subsection{On the Number of Classes}
\begin{table}[]
\centering
\resizebox{0.9\columnwidth}{!}{%
\begin{tabular}{ccccc}
\hline
\textbf{Bit-width}    & \textbf{\# Classes} & \textbf{RN18}  & \textbf{RN50}  & \textbf{MBv2}  \\ \hline
\multirow{3}{*}{W2A4} & 2 class             & \textbf{65.90} & \textbf{70.78} & \textbf{55.08} \\
                      & 3 class             & 65.74          & 70.55          & 54.51          \\
                      & 4 class             & 65.89          & 70.65          & 54.57          \\ \hline
\end{tabular}%
}
\caption{Effect of the number of classes in prompt on the accuracy of the quantized model.}
\label{tab:prompting-class}
\vspace{-3mm}
\end{table}
To examine how the number of classes included in a \textbf{mixup-prompt class} affects model accuracy, we conducted experiments varying the number of classes per text prompt (2, 3, and 4) under W2A4 quantization. 
The results, summarized in Table~\ref{tab:prompting-class}, indicate that using more classes does not necessarily lead to better performance;
the two-class case consistently achieved the highest accuracy across all three models. 
For instance, ResNet18 achieved 65.90\% and MobileNetV2 achieved 55.08\% with two classes, both outperforming the three-class and four-class configurations. 
A similar trend was observed for ResNet50, where the two-class prompt yielded 70.78\% accuracy, higher than the three-class and four-class.

These findings suggest that including too many class semantics within a single prompt could dilute the clarity of the text conditions, leading to less effective augmentation. 
In contrast, combining two classes appears to offer a better trade-off between diversity and semantic coherence, resulting in more effective data augmentation and improved model robustness.
\section{Conclusion}
\label{sec:majhead}

In this paper, we propose a simple yet effective image synthesis method that leverages text prompts to enhance the diversity of synthetic data. 
Rather than relying on naive data augmentations, our approach utilizes the image generation capabilities of Stable Diffusion through mixup-based text prompting, thereby reducing generalization gap and preventing overfitting in blockwise PTQ. 
Extensive evaluations demonstrate that our proposed method achieves state-of-the-art accuracy across a variety of models, including both ViTs and CNNs.

\section{Acknowledgement}

This work was supported by the National Research Foundation of Korea(NRF) grant funded by the Korea government (MSIT) (No. RS-2025-00561961) and the Institute of Information \& communications Technology Planning \& Evaluation (IITP) grant funded by the Korea government (MSIT) (No.2022-0-00957, Distributed on-chip memory-processor model PIM (Processor in Memory) semiconductor technology development for edge applications).
{\small
\bibliographystyle{ieeenat_fullname}
\bibliography{main}
}

\end{document}